\documentclass{article}
\usepackage{ijcai21}

\usepackage{todonotes}
\usepackage{soul}

\makeatletter
 \if@todonotes@disabled
 
 \else
 
 \fi
 \makeatother

\usepackage{xargs}
\newcommandx{\chris}[2][1=]{\todo[inline, backgroundcolor=red!25,bordercolor=red,#1]{#2 -CL}}

\usepackage{lipsum}

\newcommand\blfootnote[1]{%
  \begingroup
  \renewcommand\thefootnote{}\footnote{#1}%
  \addtocounter{footnote}{-1}%
  \endgroup
}

\usepackage{times}
\usepackage{soul}
\usepackage{url}
\usepackage[hidelinks]{hyperref}
\usepackage[utf8]{inputenc}
\usepackage[small]{caption}
\usepackage{graphicx}
\usepackage{amsmath}
\usepackage{amsthm, amssymb, mathtools}
\usepackage{booktabs}
\usepackage{algorithm}
\usepackage{algorithmic}
\usepackage{multirow}

\title{A Mixed Integer Programming Approach for Verifying Properties of \\ Binarized Neural Networks}

\author{
Christopher Lazarus\footnote{Contact Author}
\And
Mykel J. Kochenderfer
\affiliations
Stanford University\\
\emails
\{clazarus, mykel\}@stanford.edu
}

\begin{document}

\maketitle

\begin{abstract}
Many approaches for verifying input-output properties of neural networks have been proposed recently. However, existing algorithms do not scale well to large networks.
Recent work in the field of model compression studied binarized neural networks (BNNs), whose parameters and activations are binary. 
BNNs tend to exhibit a slight decrease in performance compared to their full-precision counterparts, but they can be easier to verify. 
This paper proposes a simple mixed integer programming formulation for BNN verification that leverages network structure. 
We demonstrate our approach by verifying properties of BNNs trained on the MNIST dataset and an aircraft collision avoidance controller. 
We compare the runtime of our approach against state-of-the-art verification algorithms for full-precision neural networks. 
The results suggest that the difficulty of training BNNs might be worth the reduction in runtime achieved by our verification algorithm.
\end{abstract}

\section{Introduction}
\label{sec:introduction}
\blfootnote{Copyright © 2021 for this paper by its authors. Use permitted under Creative Commons License Attribution 4.0 International (CC BY 4.0).}



Neural networks have been shown to be susceptible to adversarial attacks \cite{papernot2016limitations} often leading to drastically different outputs when slightly perturbing their input which can be costly or dangerous.
Multiple approaches to evaluate the robustness of networks to adversarial attacks have been developed. 
Many of these only provide statistical assessments and focus on evaluating the network on a large but finite collection of points in the input space. 
However, given that the input space is, in principle, infinite in cardinality, it is not viable to assess the output for all the points in the input space.

Recently, new approaches have emerged as an alternative to formally certify the input-output properties of neural networks \cite{OPT-035}. 
Properties are often specified in the form of a statement that if the input belongs to a set $\mathcal X$, then the output is in a set $\mathcal Y$. 
In the context of control, this formulation can be used to verify that the network satisfies safety constraints; in the classification setting, this formulation can be used to certify that points near a training sample are assigned the same label as that sample.

Verification algorithms are typically designed to be sound, which means that they will only conclude a property holds if the property truly holds \cite{katz2017reluplex,katz2019marabou}. 
With the aim to improve efficiency, some approaches sacrifice completeness, meaning that even if a property holds, the algorithm might fail to identify it. 
Incomplete algorithms often rely on over-approximation, allowing them to scale to problems involving larger networks, high dimensional input spaces, or high dimensional output spaces.

Even when restricting the class of networks to those with Rectified Linear Unit (ReLU) activation functions (or even a more general piecewise-linear function) the problem has been shown to be NP-hard \cite{katz2017reluplex}.
The hardness of the verification problem has motivated many different approaches \cite{OPT-035}, including reachability, optimization and search algorithms. Even  incomplete algorithms struggle to verify properties of networks with sizes commonly encountered in contemporary applications. 
In principle, simpler models should be easier to verify and binarized neural networks (BNNs) \cite{bnn-ste} are simpler than traditional full precision neural networks. Their parameters are binary and their activations are binary.

Binarization is an extreme quantization scheme that can significantly reduce memory and computation requirements. 
However, binarization introduces non-differentiable and even non-continuous blocks to the computational graph of a network, which complicates the optimization used to train the network. 
However, recent work motivated by their applicability in highly constrained environments such as edge devices has enabled them to achieve performance comparable to traditional full precision networks \cite{bnn-ste}.

The reduced memory requirement and simplified computation resulting from this representation has a drawback: binarized neural networks are harder to train. A major challenge is back propagating the gradient of the weights through sign functions. There are workarounds \cite{bnn-survey} such as using straight-through estimators (STE) \cite{bnn-ste}.


This paper presents an approach to verify properties of BNNs that leverages their structure. 
The verification problem is formulated as a mixed integer programming (MIP) problem that encodes the input set $\mathcal X$ by constraining variables associated with the input layer of the network and the output set $\mathcal Y$ by constraining variables associated to the output layer of the network. 
In our approach, we leverage the binary nature of both the parameters and the activations of the network.

Experimentally, we show that our approach can verify properties of BNNs. Section \ref{sec:experiments} demonstrates the capabilities of our approach by formally assessing the adversarial robustness of BNNs trained on the MNIST \cite{lecun-mnisthandwrittendigit-2010} dataset and properties of a collision avoidance system \cite{katz2017reluplex}. 
We compare the runtime of our approach to that of the state-of-the-art implementation of a full precision network verification algorithm for the equivalent full precision networks.
Our proposed approach reduces significantly verification runtime.

\section{Background}
\subsection{Neural Networks}
Consider $F$ a feedforward neural network with $n$ layers with input $x \in D_x \subseteq \mathbb{R}^{k_0}$ and output $y \in D_y \subseteq \mathbb{R}^{k_n}$. 
Here, $y = F(x)$ corresponds to evaluating the network on input $x$ and obtaining output $y$, $k_0$ is the dimensionality of $x$, and $k_n$ is the dimensionality of $y$. 
We assume that all inputs and outputs are flattened to vectors. 
Each layer in $F$ is a function $f_i: \mathbb{R}^{k_{i-1}} \rightarrow \mathbb{R}^{k_i}$, where $k_i$ is the dimensionality of the hidden variable $z_i$ in layer $i$. Accordingly, $z_0 = x$ and $z_n = y$. 
The network can be represented by
\begin{align}
	F = f_n \circ f_n \circ f_{n-1} \circ \cdots \circ f_1
\end{align}
and 
\begin{align}
	z_i = f_i (z_{i-1}) = \sigma_i \left ( W_i z_{i-1} + b_i \right )
\end{align}
where $W_i \in \mathbb{R}^{{k_i} \times k_{i-1}}$ is the weight matrix,
$b_i \in \mathbb{R}^{k_i}$ is the bias vector, and $\sigma_i: \mathbb{R}^{k_i} \rightarrow \mathbb{R}^{k_i}$ is the activation function.

Let $z_{i,j}$ be the value of the $j$th node in the $i$th layer, $w_{i,j} \in \mathbb{R}^{1 \times k_{i-1}}$ be the $j$th row of $W_i$, and $w_{i,j,k}$ be the $k$th entry in $w_{i,j}$. Given that activation functions are usually component-wise vector functions we have that
\begin{align}
	z_{i,j} = \sigma_{i,j} \left ( w_{i,j} z_{i-1} \right ) &= \sigma \left ( \sum_k w_{i,j,k} z_{i-1,k} \right ) \\&= \sigma_{i,j} \left ( \hat{z}_{i,j} \right )
\end{align}
where $\hat{z}_i \coloneqq W_i z_{i-1}$. We dropped the bias terms in this analysis for compactness and without loss of generality.

\subsection{Binarized Neural Networks}
A binarized neural network is a network involving binary weights and activations \cite{bnn-ste}. The goal of network binarizaion is to represent the floating-point weights $W$ and the activations $z_{i,j}$ for a given layer using $1$ bit. 
The parameters are represented by:

\begin{align}
	Q(W) = \alpha B_{W} && Q(z) = \beta B_{z}
\end{align}
where $B_{W}$ and $B_{z}$ are binarized weights and binarized activations, with scaling factors $\alpha$ and $\beta$ used for batch normalization.
The $\operatorname{sign}$ function is often used to compute $Q_{W}$ and $Q_{z}$
\begin{equation}
	\operatorname{sign} (x) = \left\{\begin{array}{ll}
+1,  &\text{if } x \geq 0 \\ 
-1,  &\text{otherwise }
\end{array}\right.
\end{equation}

The above representation enables an easy implementation of batch normalization while keeping most parameters and operations binary.

In this context, the arithmetic operations needed for a forward pass of a layer $z^b$ in a binarized network $F^b$ can be computed as:
\begin{align}
	z^b_i &= \sigma \left (Q(W) z^b_i \right ) = f_i (z^b_{i-1}) = \sigma_i \left ( Q(W)_i z^b_{i-1} \right )\\
	&= \sigma \left ( \alpha \beta B_W \circledcirc B_z \right ) = \alpha \beta \operatorname{sign}\left ( B_W \circledcirc B_z \right ) 
\end{align}
where $\circledcirc$ denotes the inner product for binary vectors with bitwise operation XNOR-Bitcount.

\paragraph{Linear.}
\begin{align}
	\hat{z}_i = Q_i z_{i-1}
\end{align}
where $Q_i \in \{-1, 1\}^{k_{i+1} \times k}$.
\paragraph{Batch Normalization.}
\begin{align}
	\hat{z}_i = \alpha_{k_i} \left ( \frac{y_i - \mu_{k_i}}{\sigma_{k_i}} \right ) + \gamma_k
\end{align}
where $y = (y_1, \dots, y_{n_{k+1}})$ and $\alpha_k, \gamma_k, \sigma_{k_i} \in \mathbb{R}$.
\paragraph{Activation Function.}
\begin{align}
	z_{i} = \operatorname{sign}(\hat{z}_i)
\end{align}
where $\hat{z}_i \in \mathbb{R}^{k_{i+1}}$ and $z_{i} \in \{ -1, 1\}^{k_{i+1}}$.
\\ \\
Some BNN architectures call for binarized inputs as well. 
Our verification approach does not require binarized inputs, but this requirement is easy to incorporate by representing floating or fixed point inputs as a combination of binary inputs, either by quantizing or directly using their binary representation. 

Section \ref{sec:bnnverif} derives how each type of block can be encoded with linear constraints, enabling the formulation of a mixed-integer programming problem for verification purposes. The last layer of a network can be used in different ways, often using a softmax or an argmax operator. 
In either case, we can encode desired properties at the output of the layer before such functions, at the so called logits.

\subsection{Neural Network Verification}
\label{sec:verification}
The verification problem consists of checking whether input-output relationships of a function hold \cite{OPT-035}. A subset of the input space $\mathcal{X} \subseteq D_x$ and a subset of the output space $\mathcal Y \subseteq D_y$ are defined. In its most general form, solving the verification problem requires certifying whether the following holds:
\begin{align}
	x \in \mathcal{X} \implies y = F(x) \in \mathcal{Y}.
\end{align}
In general, the input and output sets could have any geometry but are often constrained by the verification algorithm. 
Our approach restricts the class of $\mathcal{X}$ and $\mathcal{Y}$ to closed polytopes, corresponding to the intersection of half-spaces. These regions can be encoded as a conjunction of linear constraints, which is required for our mixed integer programming formulation.

\paragraph{Applications} 
Given that neural networks provide the state-of-the-art performance for many tasks, such as perception and control \cite{katz2017reluplex} tasks,  studying their robustness has attracted significant attention \cite{papernot2016limitations}. In the context of image classification, a network $F$ assigns a value to each of the possible labels in its training set and the maximum value $\arg \max_i y_i$ is often used to impute the label of an input $x$. 
Consider an input $x_0$ with label $i^* \in \{1,\dots,l_n\}$. It would be desirable that $y_{i^*} > y_k$ for all $j \neq i^*$, which can be encoded with the following sets:
\begin{align}
	\mathcal{X} &= \left \{ x \in D_x : \left \| x-x_0  \right \|_p \leq \epsilon \right \},\label{eq:robustness1}\\
	\mathcal{Y} &= \left \{ y \in D_y : y_{i^*} > y_j \forall j \neq i^* \right \},
\label{eq:robustness2}
\end{align}
where $\epsilon$ is the radius of the allowable perturbation in the input. If $p=1$ or $p= \infty$, we have linear constraints.
Encoding the output set $\mathcal{Y}$ is not possible with a single linear program given that the maximum operator requires a disjunction of half-spaces. With our MIP formulation, the set can be encoded directly.

\subsubsection{Full Precision Neural Network Verification}
There are plenty of approaches to verifying properties of neural networks \cite{OPT-035}.
Some methods approach verification as an optimization problem, which is an idea that we will use here. 
Many methods that only work for piecewise linear functions implement a search strategy over the activation state of the nodes in the network. 
Some apply Satisfiability Modulo Theory (SMT) by iteratively searching for assignments that satisfy all given constraints while treating the non-linear constraints lazily. 
One such method is the Reluplex algorithm \cite{katz2017reluplex,katz2019marabou}, which can be used to verify properties of binarized neural networks. However, without significant modification, it would not exploit the specific characteristics of BNNs.

\subsubsection{Binarized Neural Network Verification}
There are a few approaches designed specifically for verifying properties of BNNs. Some approaches rely on SAT solvers by reducing the verification problem to a Boolean satisfiability problem \cite{narodytska2017verifying,Narodytska2020In} which limits their applicability exclusively to fully binarized networks (networks with exclusively binary parameters and activations). 
Recently, a SAT solver based approach that is able to handle BNN constraints to speedup the verification process was introduced \cite{kai}.

Another approach that can be applied to networks with both binary and full precision parameters and piece-wise linear activations functions was recently introduced \cite{amir2021smtbased} and is an SMT-based technique that extends the Reluplex algorithm \cite{katz2019marabou} and includes various optimizations specifically designed to leverage BNNs.

Our approach relies on a simple mixed integer linear programming formulation, allowing us to handle both fully and partially binarized neural networks. 
A similar approach has been demonstrated to work well for DNNs \cite{tjeng2018evaluating} and MIP has been also used for reachability analysis \cite{lalit-ilp}. Our main contribution is to apply MIP to the verification of BNNs.

\section{Verification of BNNs using Mixed Integer Programing}
\label{sec:bnnverif}

Binarized neural networks are composed of piecewise-linear layers that may be fully connected, convolutional, or average-pooling that may have piecewise-linear activation functions, such as ReLU, $\max$ and $\operatorname{sign}$. Other commonly used elements such as batch normalization, shortcut connections, and dropout can be also characterized as affine transformations at evaluation time.

In order to address the verification problem in Section \ref{sec:verification}, we encode each of the components of the network as a set of linear constraints and then apply existing MIP technology to solve it. We encode the input and the linear equations that describe the forward-pass of the network. 
We then encode $\mathcal{Y}^C$, the complement of the output set $\mathcal{Y}$, and search for a feasible assignment. Any feasible assignment corresponds to a counterexample of the desired property. If no feasible assignment is found, then we can conclude that the network maps all the points in $\mathcal{X}$ to $\mathcal{Y}$, which certifies the property.

Below we present the formulation of each block as a set of linear constraints.
\paragraph{Input and Output Sets.}
The verification problem consists of determining whether all the elements in an input set $\mathcal{X}$ map to an output set $\mathcal{Y}$. In order to formulate that problem as a MIP, then $\mathcal{X}$ must be expressible as a conjunction of linear constraints. Additionally, given that the MIP solver will search for a feasible assignment, we need to formulate $\mathcal{Y}^C$ as another polyhedron given that
\begin{align}
	x \in \mathcal{X} \implies &y = F(x) \in \mathcal{Y}\\
	&\iff \\
	x \in \mathcal{X} \implies &y =F(x) \notin \mathcal{Y}^c.
\end{align}
Therefore, the input set and the complement of the output set have to be encoded as linear constraints on their corresponding variables.

\paragraph{Linear.} $\hat{z}_i = Q_i z_{i-1}$ where $Q_i \in \{-1,1\}^{k_{i+1} \times k_i} $
\begin{align}
	\hat{z}_{i,j} = q_j^T z_{i-1} && j=1,\dots, k_{i+1}
\end{align}

\paragraph{ReLU.} $z_i = \operatorname{ReLU}(\hat{z}_i) = \max(0,\hat{z}_i)$ and given that $l \leq \hat{z}_i \leq u$ and $\beta \in \{0,1\}$, we can encode the block as:
\begin{align}
	 z_i &\leq \hat{z}_i - l (1-\beta) \\ \hat{z}_i &\leq z_i \\ z_i&\leq \beta \cdot u \\ 0 &\leq z_i \\ \beta &\in \left \{ 0,1 \right \}
\end{align}

\paragraph{Sign.} $z_i=\operatorname{sign}(\hat{z}_i)$ 
\begin{align}
	\hat{z}_i \geq 0 \implies z_i = 1\\
	\hat{z}_i < 0 \implies z_i = 0
\end{align}
but given bounds $l \leq z_i                                                                                      \leq u$, this can be formulated as
\begin{align}
	-1 &\leq z_i\\
	z_i &\leq 1\\	
	 l \cdot \beta &\leq \hat{z}_i\\ 
	 \hat{z}_i &\leq u (1-\beta) \\
	z_i &= 1-2\cdot \beta\\
	    \beta &\in \left \{ 0,1 \right \}
\end{align}
Additionally, given the structure of the linear block that precedes the $\operatorname{sign}$ block and the fact that the variables only take values in $\{-1, 1\}$, we can always easily compute $l \geq -k_i$ and $u \leq k_i$.

\paragraph{Batch Normalization.} The batch normalization blocks have the output of linear blocks as their input.
\begin{align}
	z_i = \alpha_{k_i} \left ( \frac{y_i - \mu_{k_i}}{\sigma_{k_i}} \right ) + \gamma_k
\end{align}
where $y = (y_1, \dots, y_{n_{k+1}})$ and $\alpha_k, \gamma_k, \mu_k \sigma_{k_i} \in \mathbb{R}$, assuming $\sigma_{k_i} > 0$.
which can be rewritten as
\begin{align}
\sigma_{k_i} z_i = \alpha_{k_i}y_i - \alpha_{k_i} \mu_{k_i} + \sigma_{k_i} \gamma_{k_i}&& i=1,\dots ,n_{k+1}
\end{align}
which is a linear constraint.

\textbf{Max.} $y = \max(x_1, x_2,\dots,x_m)$ and $l_i \leq x_i \leq u_i$ can be formulated as a set of linear constraints in the following way:
\begin{align}
	y \leq x_i + (1-\beta_i) (u_{\max, -i} - l_i) &&i =1,\dots,m\\
	y \geq x_i &&i=1,\dots,m\\
	\sum_{i=1}^m \beta_i = 1\\
	\beta_i \in \{0,1\}
\end{align}

In practice, BNNs are implemented by repeated compositions of the blocks described above. In the case of BNNs, the usual order of the linear layer and batch normalization are exchanged as suggested by \cite{bnn-ste} but the MIP formulation is equivalent.

Using the above formulations, we proceded to encode the verification problem by parsing a BNN to construct the constraints that represent its forward pass and the corresponding input and output set constraints. We then call a mixed integer program solver and search for a feasible solution. If a feasible solution is found, then it certifies that the property does not hold and it serves as a counter example. If the search for a feasible solution terminates, given the completeness of the procedure, we certify that the property holds.

We implemented a tool that parses networks and produces their corresponding encoding for the solver. In the following section, we present the experiments we used to evaluate our approach and compare it to traditional DNN verification tools.
\section{Experiments}
\label{sec:experiments}
To demonstrate our approach, we performed a series of experiments. We then compared the performance of our approach to that of other publicly available verification algorithms. However, given the fact that our algorithm is specifically designed for BNNs, we decided to compare its performance to the performance of verification algorithms for DNNs. Even though the networks are not identical, we opted for this  comparison to motivate the use of BNNs for tasks where DNNs are normally favored given that we expect BNN verification to be significantly faster.

\subsection{MNIST}
\label{sec:experiments-MNIST}
We trained DNNs and BNNs on the MNIST dataset \cite{lecun-mnisthandwrittendigit-2010}. We did not tune hyper-parameters, and we used a very simple training setup. The test set accuracy of the DNN was $98.2\%$ and the BNN $95.6\%$. We thresholded all the grayscale values to black and white, which is equivalent to adding a binarization block at the input of the network.

We verified robustness properties using the input and output sets defined in equations \ref{eq:robustness1} and \ref{eq:robustness2} by allowing a maximum perturbation of $\epsilon$ around known input points using the $p=\infty$ norm. We set a time limit of $120$ seconds for each property and report our results in Table \ref{table:mnist}.

\begin{table}[h]
\centering
\begin{tabular}{@{}lrrrrrr@{}}
\toprule
\multicolumn{1}{c}{} & \multicolumn{1}{l}{$\epsilon$} & \multicolumn{3}{c}{Time (s)}                                                 & \multicolumn{2}{c}{Accuracy (\%)}                            \\ \midrule
\multicolumn{1}{c}{} & \multicolumn{1}{l}{}           & \multicolumn{1}{c}{Mean} & \multicolumn{1}{c}{Max} & \multicolumn{1}{c}{timeout} & \multicolumn{1}{c}{Verified} & \multicolumn{1}{c}{Data} \\ \midrule
BNN                  & 0.1                             & 0.223                     & 3.21                    & 0.00                    & 88.24                        & 95.6                     \\
DNN                  &                            & 5.47                     & 28.12                   & 0.05\%                  & 94.33                        & 98.22                    \\ \midrule
BNN                  & 0.3                             & 0.194                     & 4.54                    & 0.00                    & 61.78                        & 95.6                     \\
DNN                  &                             & 7.12                     & 41.33                   & 1.02\%                  & 80.68                        & 98.22                    \\ \bottomrule
\end{tabular}
\caption{MNIST results. The $\epsilon$ column indicates the maximum allowed perturbation that defines $\mathcal{X}$, The Mean column corresponds to the average time needed for the properties that did not timeout, the Max column shows the maximum time taken to verify a property that did not time out, the timeout column shows the proportion of properties that exceeded the time limit. The verified accuracy corresponds to the proportion of samples that for which the input set $\mathcal{X}$ defined a property that was verified, whereas the data accuracy columns shows the accuracy of the network evaluated on the test set.}
\label{table:mnist}
\end{table}


\subsection{ACAS}
\label{sec:experiments-ACAS}
We used a networks and property introduced by Katz et al. \shortcite{katz2017reluplex}. We trained a BNN version of the ACAS controller and tested a subset of Property 10 that only requires running one query. We sliced the state space by setting the time until loss of vertical separation $\tau = 5$ and the previous advisory as clear-of-conflict. We fixed the speed of ownship and speed of intruder values to simplify the property. For the BNN, we used 8-bits to represent each input and added a layer at the input to decrease the dimensionality to that of the original ACAS network. Quantizing the input can significantly alter the behavior of the controller, our goal was simply to assess the speedup of verification. We also trained versions of the network with full precision inputs but neither satisfied the property. We report our results in Table \ref{table:acas}.

\begin{table}[]
\centering
\begin{tabular}{@{}llrrr@{}}
\toprule
                                & \multicolumn{1}{c}{} & \multicolumn{1}{l}{Loss} & \multicolumn{1}{l}{time (s)} & \multicolumn{1}{l}{result} \\ \midrule
\multirow{3}{*}{full precision} & BNN                  & 2174.43                  & 2.37                         & violated                   \\
                                & DNN                  & 1203.44                  & 41.44                        & holds                      \\ \midrule
\multirow{1}{*}{8 bit}          & BNN                  & 1634.25                  & 5.73                         & holds                      
                                \\ \bottomrule
\end{tabular}
\caption{ACAS results.}
\label{table:acas}
\end{table}

\section{Conclusion}

Our results indicate that our simple MIP approach for verifying properties of BNNs performs significantly faster than other methods for DNNs. Training BNNs is challenging but the reduction in verification cost should be considered and incentivize their use for safety-critical applications that require verification of certain properties.

In our MNIST experiments we were able to verify the robustness of networks with about a $10\times$ reduction in verification time.

Our ACAS experiments show that our approach is able to verify BNNs that implement controllers in about $20\times$ reduced time. BNNs appear to be particularly well suited as controllers for safety critical systems.

Our proposed approach encodes BNN as mixed integer linear programs and is able to verify properties of binarized neural networks and partially binarized neural networks. 
Our experiments indicate that this approach is about an order of magnitude faster than verifying properties of comparable full precision neural networks.

The ease with which we can verify BNNs should increase their use for safety critical applications. 
BNNs are harder to train, but the difficulty might be worth the cost given how much faster verification becomes along with the efficient hardware implementations that they enable. 
The use of BNNs has some drawbacks and requires considerations such as how to handle non-binary inputs. 
Quantizing the inputs allows us to preserve the binary architecture but decreases the applicability of BNNs because some applications might have continuous input domains that would be better modeled with floating point numbers. 
However, it appears that even verified networks with floating point parameters are potentially unsafe \cite{fperror}.

We used a general purpose mixed integer programming solver. A potential area of future research would be to design a MIP solver that exploits some of the BNN specific characteristics of the problems to further decrease verification time. Another potential direction would be to train ternary BNNs in order to explore the impact of sparsification on runtime, given that he equations that encode ternary BNNs require fewer variables





\bibliography{binverif-ijcai}

\begin{thebibliography}{}

\bibitem[\protect\citeauthoryear{Amir \bgroup \em et al.\egroup
  }{2021}]{amir2021smtbased}
Guy Amir, Haoze Wu, Clark Barrett, and Guy Katz.
\newblock An {SMT}-based approach for verifying binarized neural networks.
\newblock {\em Tools and Algorithms for the Construction and Analysis of
  Systems, TACAS}, 2021.

\bibitem[\protect\citeauthoryear{Hubara \bgroup \em et al.\egroup
  }{2016}]{bnn-ste}
Itay Hubara, Matthieu Courbariaux, Daniel Soudry, Ran El-Yaniv, and Yoshua
  Bengio.
\newblock Binarized neural networks.
\newblock In {\em Advances in Neural Information Processing Systems}, 2016.

\bibitem[\protect\citeauthoryear{Jia and Rinard}{2020a}]{kai}
Kai Jia and Martin Rinard.
\newblock Efficient exact verification of binarized neural networks.
\newblock In {\em Advances in Neural Information Processing Systems}, 2020.

\bibitem[\protect\citeauthoryear{Jia and Rinard}{2020b}]{fperror}
Kai Jia and Martin Rinard.
\newblock Exploiting verified neural networks via floating point numerical
  error.
\newblock {\em CoRR}, abs/2003.03021, 2020.

\bibitem[\protect\citeauthoryear{Katz \bgroup \em et al.\egroup
  }{2017}]{katz2017reluplex}
Guy Katz, Clark Barrett, David~L. Dill, Kyle Julian, and Mykel~J Kochenderfer.
\newblock Reluplex: An efficient {SMT} solver for verifying deep neural
  networks.
\newblock In {\em International Conference on Computer Aided Verification},
  2017.

\bibitem[\protect\citeauthoryear{Katz \bgroup \em et al.\egroup
  }{2019}]{katz2019marabou}
Guy Katz, Derek~A Huang, Duligur Ibeling, Kyle Julian, Christopher Lazarus,
  Rachel Lim, Parth Shah, Shantanu Thakoor, Haoze Wu, Aleksandar Zelji{\'c},
  et~al.
\newblock The {M}arabou framework for verification and analysis of deep neural
  networks.
\newblock In {\em International Conference on Computer Aided Verification},
  2019.

\bibitem[\protect\citeauthoryear{LeCun and
  Cortes}{2010}]{lecun-mnisthandwrittendigit-2010}
Yann LeCun and Corinna Cortes.
\newblock {MNIST} handwritten digit database.
\newblock 2010.

\bibitem[\protect\citeauthoryear{Liu \bgroup \em et al.\egroup
  }{2021}]{OPT-035}
Changliu Liu, Tomer Arnon, Christopher Lazarus, Christopher Strong, Clark
  Barrett, and Mykel~J. Kochenderfer.
\newblock Algorithms for verifying deep neural networks.
\newblock {\em Foundations and Trends® in Optimization}, 4(3-4):244--404,
  2021.

\bibitem[\protect\citeauthoryear{Lomuscio and Maganti}{2017}]{lalit-ilp}
Alessio Lomuscio and Lalit Maganti.
\newblock An approach to reachability analysis for feed-forward relu neural
  networks.
\newblock {\em CoRR}, abs/1706.07351, 2017.

\bibitem[\protect\citeauthoryear{Narodytska \bgroup \em et al.\egroup
  }{2018}]{narodytska2017verifying}
Nina Narodytska, Shiva Kasiviswanathan, Leonid Ryzhyk, Mooly Sagiv, and Toby
  Walsh.
\newblock Verifying properties of binarized deep neural networks.
\newblock In {\em AAAI Conference on Artificial Intelligence (AAAI)}, 2018.

\bibitem[\protect\citeauthoryear{Narodytska \bgroup \em et al.\egroup
  }{2020}]{Narodytska2020In}
Nina Narodytska, Hongce Zhang, Aarti Gupta, and Toby Walsh.
\newblock In search for a {SAT}-friendly binarized neural network architecture.
\newblock In {\em International Conference on Learning Representations}, 2020.

\bibitem[\protect\citeauthoryear{Papernot \bgroup \em et al.\egroup
  }{2016}]{papernot2016limitations}
Nicolas Papernot, Patrick McDaniel, Somesh Jha, Matt Fredrikson, Z~Berkay
  Celik, and Ananthram Swami.
\newblock The limitations of deep learning in adversarial settings.
\newblock In {\em IEEE European Symposium on Security and Privacy (EuroS\&P)},
  pages 372--387. IEEE, 2016.

\bibitem[\protect\citeauthoryear{Simons and Lee}{2019}]{bnn-survey}
Taylor Simons and Dah-Jye Lee.
\newblock A review of binarized neural networks.
\newblock {\em Electronics}, 8(6):4--5, 2019.

\bibitem[\protect\citeauthoryear{Tjeng \bgroup \em et al.\egroup
  }{2019}]{tjeng2018evaluating}
Vincent Tjeng, Kai~Y. Xiao, and Russ Tedrake.
\newblock Evaluating robustness of neural networks with mixed integer
  programming.
\newblock In {\em International Conference on Learning Representations}, 2019.

\end{thebibliography}
\bibliographystyle{named}

\end{document}